\documentclass[runningheads]{llncs}

\usepackage[utf8]{inputenc}
\usepackage{graphicx}
\usepackage{subcaption}
\usepackage[skip=1ex,font=small,labelsep=period]{caption}
\usepackage{amsmath,amssymb} 
\usepackage{color}
\usepackage[pagebackref=true,breaklinks=true,letterpaper=true,colorlinks=true,urlcolor=bblue,bookmarks=false,allcolors=black]{hyperref}

\newcommand{\mytilde}{\raise.17ex\hbox{$\scriptstyle\sim$}}

\newcommand{\eg}{e.g.~}

\begin{document}

\title{A Summary of the 4th International Workshop on~Recovering 6D Object Pose} 

\titlerunning{4th International Workshop on~Recovering 6D Object Pose}

\newcommand{\namesep}{, }
\author{
Tomáš~Hodaň\inst{1}\namesep
Rigas~Kouskouridas\inst{2}\namesep
Tae-Kyun~Kim\inst{3}\namesep
Federico~Tombari\inst{4}
Kostas~Bekris\inst{5}\namesep
Bertram~Drost\inst{6}\namesep
Thibault~Groueix\inst{7}\namesep
Krzysztof~Walas\inst{8}
Vincent~Lepetit\inst{9}\namesep
Ales~Leonardis\inst{10}\namesep
Carsten~Steger\inst{6}\namesep
Frank~Michel\inst{11}
Caner~Sahin\inst{3}\namesep
Carsten~Rother\inst{12}\namesep
Jiří~Matas\inst{1}
}

\authorrunning{Hodaň, Kouskouridas, Kim et al.}

\institute{\small{
 $^{1}$CTU~in~Prague\namesep
 $^{2}$Scape Technologies\namesep
 $^{3}$Imperial~College~London\namesep
 $^{4}$TU~Munich
 $^{5}$Rutgers University\namesep
 $^{6}$MVTec\namesep
 $^{7}$Ecole~Nationale~des~Ponts~et~Chaussées
 $^{8}$Poznan~University~of~Technology\namesep
 $^{9}$University~of~Bordeaux
 $^{10}$University~of~Birmingham\namesep
 $^{11}$TU~Dresden\namesep
 $^{12}$Heidelberg~University
}}

\maketitle

\begin{abstract}

This document summarizes the 4th International Workshop on Recovering 6D Object Pose which was organized in conjunction with ECCV 2018 in Munich. The workshop featured four invited talks, oral and poster presentations of accepted workshop papers, and an introduction of the BOP benchmark for 6D object pose estimation. The workshop was attended by 100+ people working on relevant topics in both academia and industry who shared up-to-date advances and discussed open problems.

\end{abstract}

\section{Introduction}

An accurate, fast and robust estimation of 6D object pose is of great importance to many application fields such as robotic manipulation, augmented reality, scene interpretation, and autonomous driving. In robotics, for example, the 6D object pose facilitates spatial reasoning and allows an end-effector to act upon the object. In an augmented reality scenario, object pose can be used to enhance one’s perception of reality by augmenting objects with extra information such as hints for assembly.

The introduction of consumer and industrial grade RGB-D sensors have allowed for substantial improvement in 6D object pose estimation as these sensors concurrently capture both the appearance and geometry of the scene. However, there still remain challenges to be addressed, including robustness against occlusion and clutter, scalability to multiple objects, and fast and reliable object modelling, including capturing of reflectance properties. Extending contemporary methods to work reliably and with sufficient execution speed in an industrial setting is still an open problem. Many recent methods focus on specific rigid objects, but pose estimation of deformable or articulated objects and of object categories is also an important research direction.

\begin{figure}[!t]
\begin{center}
    \includegraphics[width=1.0\columnwidth]{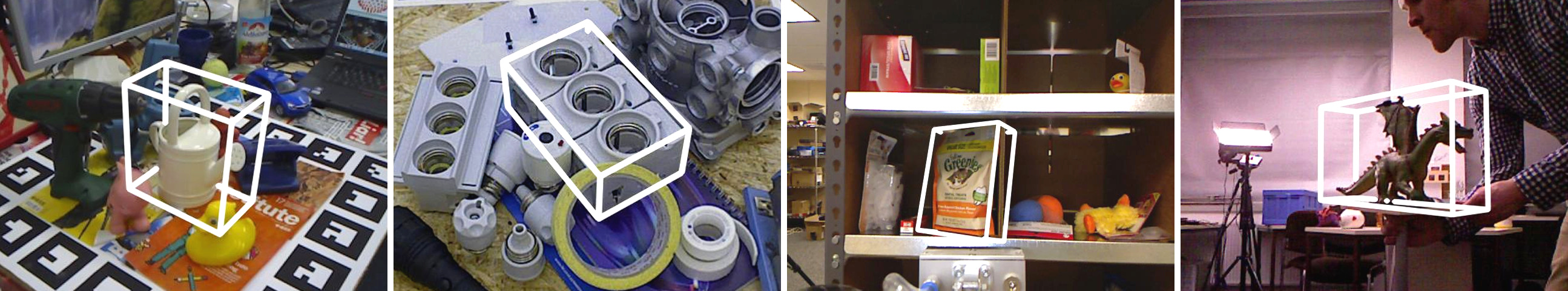}
    \caption{\label{fig:6d} Estimation of 6D pose, i.e. 3D translation and 3D rotation, of a specific rigid object. This task is considered in the BOP benchmark~\cite{hodan2018bop}.}
\end{center}
\end{figure}

The field of 6D object pose estimation has gained more attention last years. A big achievement for the field is the best paper award of ECCV 2018 given to Martin Sundermeyer, Zoltan Marton, Maximilian Durner, Manuel Brucker, and Rudolph Triebel for their work titled \emph{Implicit 3D Orientation Learning for 6D Object Detection from RGB Images}.

The 4th edition of the International Workshop on Recovering 6D Object Pose~\cite{r6d2018}\footnote{The previous editions were held at ICCV 2015~\cite{r6d2015}, ECCV 2016~\cite{r6d2016}, and ICCV 2017~\cite{r6d2017}.} was organized in conjunction with ECCV 2018 and was attended by 100+ people from both academia and industry who shared up-to-date advances and discussed open problems. Four invited speakers talked about their current work (Section~\ref{sec:invited_talks}), the BOP benchmark for 6D object pose estimation was introduced (Section~\ref{sec:bop}), and the accepted workshop papers were presented.

The workshop covered the following topics: (a)~6D object pose estimation (a.k.a. 3D object detection) and tracking, (b)~3D object modeling and reconstruction, (c)~surface representation and registration, (d)~robustness to occlusion and background clutter, (e)~multiple object instance detection, (f)~pose estimation of non-rigid objects and object categories, (g)~robotic grasping and grasp affordances, and (h)~object manipulation and interaction.

Many methods for 6D pose estimation of specific rigid objects (Fig.~\ref{fig:6d}) have been published recently, but were usually compared with only a few competitors on a small subset of datasets. It had been therefore unclear which methods perform well and in which scenarios. To capture the \emph{status quo} of the field, we organized the SIXD Challenge~\cite{sixd2017} at the 3rd workshop edition held at ICCV 2017~\cite{r6d2017}. The results submitted to the challenge were published in the BOP benchmark paper~\cite{hodan2018bop} and presented at the 4th workshop edition.

Papers submitted to the workshop were peer-reviewed. Out of 13 submissions, the program committee accepted 10 papers which were introduced at the workshop through oral and poster presentations. The best paper award was given to \emph{Image-to-Voxel Model Translation with Conditional Adversarial Networks} by Vladimir Knyaz, Vladimir V. Kniaz, and Fabio Remondino.

\section{Invited talks} \label{sec:invited_talks}

The invited talks were given by Federico Tombari from Technical University of Munich, Kostas Bekris from Rutgers University, Bertram Drost from MVTec, and Thibault Groueix from Ecole Nationale des Ponts et Chaussées. The talks are summarized below and the slides are available on the workshop website~\cite{r6d2018}.

\subsection{From 3D Descriptors to Monocular 6D Pose: What Have We Learned? -- \textit{Federico Tombari}}

While 6D rigid pose estimation has been an important research task for more than twenty years, recently the design of new algorithms is more and more focused on overcoming the limitations provided by real world applications, so to bridge the gap between lab research and products. This translates to the necessity to move on from the simplified scenario of estimating the pose of a single object on a clutter-less planar surface, towards scenarios with high clutter and occlusion~\cite{tan2017looking}. At the same time, algorithms need to process input data at a very high frame rate, so to reduce the latency of the output and avoid lagging, provide accurate pose estimation (e.g. to reduce jitter during tracking) and cope with resource-limited hardware architectures such as mobile phones or embedded computers. This is particularly motivated by applications that already have a strong market interest such as augmented reality, personal/industrial robotics and autonomous driving. A disruptive technology that strongly influenced the field of 6D rigid pose estimation in the past 5 years is deep learning. Hence, the talk presented an overview of the current trends and results regarding the use of deep learning for this task, in view of overcoming the aforementioned limitations. In particular, it highlighted two main research directions where deep learning has been leveraged to improve the state of the art: i) the definition of 3D descriptors for 3D data; ii) 6D object pose estimation from RGB (or monocular) data.  

As for the first aspect, we briefly went over the development of 3D descriptors for unorganized 3D representations, starting from the handcrafted ones \cite{tombari10SHOT,rusu09FPFH} until the more recent "learned" ones. An important aspect regarding the influence of deep learning on this field is that certain 3D representations such as point clouds and 3D meshes, frequently utilized for 6D rigid pose estimation tasks, are not well suited to convolutions due to their intrinsic unorganized nature. Hence, an important step in the direction of learning 3D features was the introduction of methods such as PointNet \cite{qi2017pointnet} and \cite{wang18DGCNN}, that, conversely to approaches such as 3D Match \cite{zeng173DMatch} and \cite{khoury17CGN}, can directly operate on point clouds or meshes without the need to either voxelize or histogram the data. The recent global and fully-convolutional architecture for point cloud processing and scene understanding proposed in \cite{rethage18} was also introduced. 

As for the second research direction, the intuition of estimating the 6D pose of an object based on monocular information relies on the fact that humans can often have a rough idea of the pose of the objects in the surrounding 3D space simply from monocular cues, provided that they are familiar with the shape of the objects. Several works have recently explored this direction, which were briefly introduced and compared in terms of characteristics. One distinctive trait that differentiates such approaches is the type of output that the network is trained to infer: from the 8 corners of the projected bounding box \cite{rad2017bb8,tekin18} to the regression of the 6D pose \cite{xiang18posecnn,do18deep6dpose} or the classification of the viewpoint and in-plane rotations \cite{kehl2017ssd}. The method in \cite{manhardt18} was also introduced, that proposes to learn the 6D pose refinement from pairs of RGB patches using a CNN. Finally, the extension of monocular 6D pose estimation to the autonomous driving domain was also discussed, by referencing recent directions such as \cite{chen16Mono3D,chen153DOP}.  

In conclusion, deep learning appears as a powerful tool for 6D rigid pose estimation, although seems still strongly limited by open issues such as generalizability, learning of geometric invariance and computational efficiency (especially in a field where most applications need to deal with resource-limited hardware). Monocular pose estimation can be promisingly carried out via deep learning, although not yet as accurately as with a depth sensor, as also showcased by a qualitative comparison on a real sequence between the two tracking approaches in \cite{manhardt18} (monocular) and \cite{tan2017looking} (RGB-D). Finally, deploying monocular 6D pose estimation jointly with monocular semantic SLAM such as CNN-SLAM \cite{tateno17CNNSLAM}, which also leverages deep learning to obtain dense semantic reconstruction, appears as an interesting direction to explore towards full (i.e. semantic+geometry) scene understanding from monocular data.

\subsection{Towards Robust 6D Pose Estimation: Physics-based Reasoning and Data Efficiency -- \textit{Kostas Bekris}}

Towards the objective of robust 6 DoF pose estimation at an accuracy
and speed level that allows robots to manipulate objects in their
surroundings, this talk focused on warehouses tasks, such as picking
from bins, packing and sorting. Warehouses can be seen as a stepping
stone between the success story of robotics in manufacturing and the
vision of deploying robots in everyday human environments.  Warehouses
involve a large variety of objects, which can appear in general,
unpredictable configurations, as well as cluttered scenes and tight
spaces, such as shelving units, which limit the observability of
on-board sensors and introduce occlusions. They allow, however, access
to known object models, which frequently correspond to standard
geometric shapes, due to packaging.

In these setups, it is critical for a robot to both utilize
physics-based reasoning to achieve 3D scene-level understanding as
well as minimize the dependence of solutions to excessive human
labeling, which negatively impacts scalability. With these priorities
in mind, the talk highlighted a pipeline for robust 6D pose
estimation, which includes the following four steps: 1) semantic object segmentation given physically realistic synthetic data, 2) pose hypothesis generation through robust point registration, 3) pose improvement via consideration of physical constraints at the scene level, and 4) lifelong self-learning through active physical interaction with objects.

Recently, deep learning methods, such as those employing Convolutional
Neural Networks (CNN's), have become popular for object
detection \cite{ren2015faster,redmon2016you} and pose estimation
\cite{kehl2017ssd,xiang2017posecnn}, outperforming alternatives in
most benchmarks. These results are typically
obtained by training CNN's using datasets with a large number of labeled images. Such datasets, however, need to be
collected in a way that captures the intricacies of the environment
the robot is deployed in, such as lighting conditions, occlusion and clutter. This motivates the development of synthetic dataset that can capture the known parameters of the environment and generate data accordingly, while avoid overfitting to
the unknown parameters.

\emph{The first component of the proposed pipeline} is to use a
physics engine in the synthetic dataset generation pipeline
\cite{mitash2017self}. The physics engine defines environmental
constraints on object placement, which naturally capture in the
training set, the distribution of object poses that can realistically
appear during testing.  Furthermore, a physics engine is a convenient
tool to parameterize the unknown scene features, such as illumination.
A randomization over such parameters is very effective in avoiding overfitting to synthetic textures of objects.

Given semantic object segmentation, the problem of estimating the 6D object
poses involves geometric reasoning regarding the position
and orientation of the detected objects. Solutions that became popular
in the context of the Amazon Picking Challenge (APC)
\cite{Correll:2016aa}, use a Convolutional Neural Network (CNN) for object segmentation \cite{Princeton,hernandez2016team}
followed by a 3D model alignment step using point cloud
registration techniques \cite{mellado2014super,icp}. The quality of
the pose estimate, however, can still suffer due to over-reliance on
the learned models.

\emph{The second insight of the talk} is that CNN output can be
seen as a probability for an object to be visible at each pixel. These
segmentation probabilities can then be used during the registration
process to achieve robust and fast pose estimation. This requires
sampling a base of points on a point cloud segment, such that all
points on the base belong to the target object with high
probability. The resulting approach, denoted as ``Stochastic Congruent
Sets'' (StoCS) \cite{Mitash:2018aa}, builds a
probabilistic graphical model given the obtained soft segmentation and
information from the pre-processed geometric object models. The
pre-processing corresponds to building a global model descriptor that
expresses oriented point pair features \cite{drost2010model}.  This
geometric modeling, not only biases the base samples to lie within the
object bound, but is also used to constrain the search for finding the
congruent sets, which provides a substantial computational benefit.

\emph{The third key observation of the talk} is to treat
individual-object predictions with some level of uncertainty and
perform a global, scene-level optimization process that takes object
interactions into account \cite{Mitash:2018ab}. This information
arises from physical properties, such as respecting gravity and
friction as well as the requirement that objects do not penetrate one
another. In particular, a Monte Carlo Tree Search (MCTS) process
utilizes local detections to achieve scene-level optimization. It
generates multiple candidate poses per object and then searches over
the cartesian product of these individual object pose candidates to
find the optimal scene hypothesis. The scene is evaluated according to
a score defined in terms of similarity of the rendered hypothesized
scenes against input data. The search performs constrained local
optimization via physics correction and ICP \cite{icp}.  Through
this physical reasoning, the resulting pose estimates for the objects
are of improved accuracy and by default consistent. 

Once the system has access to an object detector and a pose estimation
process, it can already be deployed for the desired task.
Nevertheless, as the system performs its task, it also gets access to
data in the operation domain, which it did not have access to during
training. This data could be very useful in further improving the
performance of the system but they are not labelled.

\emph{The fourth aspect of the talk} is a solution for automatically
labeling real images acquired from multiple viewpoints using a robotic
manipulator \cite{mitash2017self}.  A robotic manipulator
autonomously collects multi-view images of real scenes and labels them
automatically using the object detector trained with the above
physics-based simulation tool. The key insight is the fact that the
robot can often find a good viewing angle that allows the detector to
accurately label the object and estimate its pose. The object's
predicted pose is then used to label images of the same scene taken
from more difficult and occluded views. The transformations between
different views are known because they are obtained by moving the
robotic manipulator. Overall, the data can be added to the existing
synthetic dataset to re-train the model for better performance as part
of a lifelong, self-learning procedure.

\subsection{Detecting Geometric Primitives in 3D Data -- \textit{Bertram Drost}}

Even though methods based on (deep) learning lead to major advances in many areas of computer vision over the last years, the top-performing methods for 6D object detection are still hand-crafted, classic methods. This is evident from the recent results on the BOP benchmark (Section~\ref{sec:bop}).

Methods that currently perform best on BOP are based on a voting scheme that can be interpreted as a meet-in-the-middle between RANSAC and a generalized Hough transform.
The base method~\cite{drost2010model} uses a local parametrization for the object pose, where an oriented scene point (reference point) is fixed and assumed to be on the target object. The remaining local parameters are then the point on the model surface that corresponds to the reference point and the rotation around the normal vector, which combined represent three degrees of freedom. The optimal local parameters are recovered using a voting scheme. For this, the reference point is paired with neighboring scene points, and similar point pairs on the model are searched using an efficient hashing scheme. Each such match then casts a vote.
While the base method shows good results for 3D shapes with a distintive geometry, it has weaknesses for shapes that are symmetric of strongly self-similar. This is mostly because point pairs on such shapes are no longer very discriminative. The voting thus finds all symmetric poses simultaneously, which is both slower and less robust.

The talk presented a way of adapting the base method for geometric primitive shapes - spheres, planes, cylinders, and to some extend cones.
First, the local parameter space was reduced by removing duplicate entries due to symmetries. For example, since a sphere is identical no matter the corresponding point on its surface or the rotation around the normal, the local parameter space becomes zero-dimensional, i.e. a single counter. Additionally, the local parameters can be extended by shape parameters such as the radius of a sphere or cylinder, or the opening angle of a cone. Second, instead of using a hashing scheme for matching point pairs between scene and model, an explicit point pair matching can be used thanks to the explicit nature of the geometric primitives.
Those changes make the method both faster and more robust for such geometric primitives, while also allowing the recovery of shape parameters.

\subsection{Parameteric Estimation of 3D Surfaces and Correspondences -- \textit{Thibault Groueix}}

Pose of a rigid object has 6 degrees of freedom and is well defined.
A broader class of objects are non-rigid shapes such as articulated robots or humans. Articulated robots have an additional degree of freedom for each joint. Those additional intrinsic parameters also have to be estimated. In this case, it is easy to manually design a parametrization with N degrees of freedom, where each parameter naturally encodes the angle of a joint. Humans are much harder to parameterize and doing it manually is hard. We propose to learn the parameterization~\cite{groueix20183d}.

To that end, we want to map each shape $S$ of category $C$, represented as a point cloud, to a parameter space. A relevant prior on deformable shapes is the ability to find a neutral template, already encoding lots of general information on humans.
To learn a parameterization, we use autoencoder neural networks. The encoder is a PointNet~\cite{qi2017pointnet} and the decoder is a Shape Deformation Network~\cite{groueix20183d} that deforms the template by iteratively transforming each point on the template to a point in 3D. Given correspondence annotation between a training example and the template, the L2 distance between the generated and the input point clouds is penalized. Annotation of correspondences is expensive and can be avoided through an unsupervised reconstruction loss, the chamfer distance, and a proper regularization. The parameterization is a mapping from the parameter space to the point cloud space, and is given by the learned decoder.

At test time, the parameters associated with this parametrization can be estimated using the encoder. We propose to use this estimate as an initialization for a local exploration of the parameter space through gradient descent. The learned parametrization is evaluated on the task of 3D correspondences on the FAUST dataset~\cite{bogo2014faust} and achieves state-of-the-art-results. It also displays strong robustness to holes and noise in the data.

The method is not limited to humans, but applies to any category of shape for which a template can be found.  For broader category without natural templates such as chairs, or general furniture, the template can be replaced by a set a square patches~\cite{groueix2018atlasnet}. The assumption is that any object can be reconstruct through the deformation of a set of patches. The proposed parametrization, Atlasnet, is learned on Shapenet and applied for the task of Single View Reconstruction, leading to state-of-the-art results. This confirms that learning a parametrization can be the key to extending 6D pose estimation beyond rigid objects.

\section{BOP: Benchmark for 6D Object Pose Estimation} \label{sec:bop}

The BOP benchmark considers the task of 6D pose estimation of a rigid object from a single RGB-D input image, when the training data consists of a texture-mapped 3D object model or images of the object in known 6D poses. The benchmark comprises of: (i) eight datasets~\cite{hinterstoisser2012accv,brachmann2014learning,tejani2014latent,doumanoglou2016recovering,hodan2017tless,rennie2016dataset} in a unified format that cover different practical scenarios, including two new datasets focusing on varying lighting conditions, (ii) an evaluation methodology with a pose-error function that deals with pose ambiguities, (iii) an evaluation of 15 diverse recent methods that captures the state of the art, and (iv) an online evaluation system at \texttt{\href{http://bop.felk.cvut.cz}{bop.felk.cvut.cz}} that is open for continuous submission of new results.

The evaluation shows that methods based on point-pair features~\cite{vidal2018sixd,drost2010model}, introduced in 2010, currently perform best. They outperform template matching methods~\cite{hodan2015detection}, learning-based methods~\cite{brachmann2014learning,brachmann2016uncertainty,kehl2016deep,tejani2014latent} and methods based on 3D local features~\cite{buch2016local,buch2017rotational}.
Occlusion is a big challenge for current methods, as shown by scores dropping swiftly already at low levels of occlusion. As another important future research directions, our analysis identified robustness against object symmetries and similarities, varying lighting conditions, and noisy depth images.

\subsection{Future Plans}

Various possible extensions of the BOP benchmark regarding datasets, problem
statement and evaluation metrics were discussed at the workshop and are summarized in the following paragraphs.

The benchmark was started with the simple task of 6D localization of a single instance of a single object. This task allowed to evaluate most of the recent methods out of the box and is a common denominator of the other 6D localization variants -- a single instance of multiple objects, multiple instances of a single object, and multiple instances of multiple objects. The plan is to move step by step and add the more complicated variants to the online evaluation system in the near future, as well as the 6D detection task, where no prior information about the object presence is provided~\cite{hodan2016evaluation}. Note that while it is easy to extend the evaluation to other tasks, it is non-trivial to extend the methods.

The plan is also to keep adding new datasets, \eg \cite{drost2017introducing}, and for every new dataset to reserve a set of test images for which the ground-truth annotation will be private. The datasets that are currently included in BOP are fully public.

Another topic of discussion were the pose-error functions. In the BOP benchmark, 6D pose estimates are evaluated using the Visible Surface Discrepancy (VSD) which measures misalignment of the visible part of the object surface.
One limitation of the current version of VSD is that it does not consider color. The same holds for the other pose-error functions commonly used in the literature~\cite{hodan2016evaluation}. While the alignment of color texture is less relevant for a number of robotic applications, it may be important for augmented-reality applications. The plan is to fix this limitation in a new version of VSD with an extended pixel-wise test checking not only the distance of the object surface but also its color. 
We are open to other feedback and discussion about the evaluation methodology.

\section{Conclusions}

This document summarized the 4th International Workshop on Recovering 6D Object Pose which was organized at ECCV 2018 and featured four invited talks, oral and poster presentations of accepted workshop papers, and an introduction of the BOP benchmark for 6D object pose estimation.

An accurate, fast and robust method for 6D object pose estimation is still in need, as is evident from the current scores on the BOP benchmark~\cite{hodan2018bop}.
We would like to encourage authors of relevant methods to keep submitting results to the online evaluation system at \texttt{\href{http://bop.felk.cvut.cz}{bop.felk.cvut.cz}}. We will present snapshots of the leaderboards at the next workshop editions, which are planned for the upcoming major conferences.

\bibliographystyle{splncs04}
\bibliography{ref}

\begin{thebibliography}{10}
\providecommand{\url}[1]{\texttt{#1}}
\providecommand{\urlprefix}{URL }
\providecommand{\doi}[1]{https://doi.org/#1}

\bibitem{r6d2015}
1st {I}nternational {W}orkshop on {R}ecovering {6D} {O}bject {P}ose, {ICCV}
  2015, {S}antiago. \url{https://labicvl.github.io/3DPose-2015.html}

\bibitem{r6d2016}
2nd {I}nternational {W}orkshop on {R}ecovering {6D} {O}bject {P}ose, {ECCV}
  2016, {A}msterdam. \url{https://labicvl.github.io/R6D}

\bibitem{r6d2017}
3rd {I}nternational {W}orkshop on {R}ecovering {6D} {O}bject {P}ose, {ICCV}
  2017, {V}enice. \url{http://cmp.felk.cvut.cz/sixd/workshop\_2017/}

\bibitem{r6d2018}
4th {I}nternational {W}orkshop on {R}ecovering {6D} {O}bject {P}ose, {ECCV}
  2018, {M}unich. \url{http://cmp.felk.cvut.cz/sixd/workshop\_2018/}

\bibitem{sixd2017}
{SIXD} challenge 2017. \url{http://cmp.felk.cvut.cz/sixd/challenge\_2017/}

\bibitem{icp}
Besl, P.J., McKay, N.D.: {Method for Registration of 3D Shapes}. International
  Society for Optics and Photonics  (1992)

\bibitem{bogo2014faust}
Bogo, F., Romero, J., Loper, M., Black, M.J.: {FAUST}: Dataset and evaluation
  for 3d mesh registration. In: CVPR (2014)

\bibitem{brachmann2014learning}
Brachmann, E., Krull, A., Michel, F., Gumhold, S., Shotton, J., Rother, C.:
  Learning {6D} object pose estimation using {3D} object coordinates. In: ECCV
  (2014)

\bibitem{brachmann2016uncertainty}
Brachmann, E., Michel, F., Krull, A., Yang, M.Y., Gumhold, S., Rother, C.:
  Uncertainty-driven {6D} pose estimation of objects and scenes from a single
  {RGB} image. In: CVPR (2016)

\bibitem{buch2016local}
Buch, A.G., Petersen, H.G., Kr{\"u}ger, N.: Local shape feature fusion for
  improved matching, pose estimation and {3D} object recognition. SpringerPlus
  (2016)

\bibitem{buch2017rotational}
Buch, A.G., Kiforenko, L., Kraft, D.: Rotational subgroup voting and pose
  clustering for robust {3D} object recognition. In: ICCV (2017)

\bibitem{chen16Mono3D}
Chen, X., Kundu, K., Zhang, Z., Ma, H., Fidler, S., Urtasun, R.: Monocular 3d
  object detection for autonomous driving. In: CVPR (2016)

\bibitem{chen153DOP}
Chen, X., Kundu, K., Zhu, Y., Berneshawi, A., Ma, H., Fidler, S., Urtasun, R.:
  3d object proposals for accurate object class detection

\bibitem{Correll:2016aa}
Correll, N., Bekris, K.E., Berenson, D., Brock, O., Causo, A., Hauser, K.,
  Osada, K., Rodriguez, A., Romano, J., Wurman, P.: {Analysis and Observations
  From the First Amazon Picking Challenge}. IEEE Trans. on Automation Science
  and Engineering (T-ASE)  (2016)

\bibitem{do18deep6dpose}
Do, T.T., Cai, M., Pham, T., Reid, I.: Deep-6dpose: Recovering 6d object pose
  from a single rgb image. arXiv preprint arXiv:1802.10367  (2018)

\bibitem{doumanoglou2016recovering}
Doumanoglou, A., Kouskouridas, R., Malassiotis, S., Kim, T.K.: Recovering {6D}
  object pose and predicting next-best-view in the crowd. In: CVPR (2016)

\bibitem{drost2017introducing}
Drost, B., Ulrich, M., Bergmann, P., H{\"a}rtinger, P., Steger, C.: Introducing
  {MVT}ec {ITODD} -- a dataset for {3D} object recognition in industry. In:
  CVPR (2017)

\bibitem{drost2010model}
Drost, B., Ulrich, M., Navab, N., Ilic, S.: Model globally, match locally:
  Efficient and robust {3D} object recognition. In: CVPR (2010)

\bibitem{groueix20183d}
Groueix, T., Fisher, M., Kim, V.G., Russell, B.C., Aubry, M.: {3D-CODED}: 3d
  correspondences by deep deformation. In: ECCV (2018)

\bibitem{groueix2018atlasnet}
Groueix, T., Fisher, M., Kim, V.G., Russell, B.C., Aubry, M.: Atlasnet: A
  papier-mache approach to learning 3d surface generation. arXiv preprint
  arXiv:1802.05384  (2018)

\bibitem{hernandez2016team}
Hernandez, C., Bharatheesha, M., Ko, W., Gaiser, H., Tan, J., van Deurzen, K.,
  de~Vries, M., Van~Mil, B., van Egmond, J., Burger, R., Morariu, M., Ju, J.,
  Gerrmann, X., Ensing, R., Frankenhuyzen, J.V., Wisse, M.: Team delft's robot
  winner of the amazon picking challenge 2016. In: Robot World Cup. pp.
  613--624. Springer (2016)

\bibitem{hinterstoisser2012accv}
Hinterstoisser, S., Lepetit, V., Ilic, S., Holzer, S., Bradski, G., Konolige,
  K., Navab, N.: Model based training, detection and pose estimation of
  texture-less 3{D} objects in heavily cluttered scenes. In: ACCV (2012)

\bibitem{hodan2017tless}
Hoda{\v{n}}, T., Haluza, P., Obdr{\v{z}}{\'a}lek, {\v{S}}., Matas, J.,
  Lourakis, M., Zabulis, X.: {T-LESS}: An {RGB-D} dataset for {6D} pose
  estimation of texture-less objects. WACV  (2017)

\bibitem{hodan2016evaluation}
Hoda{\v{n}}, T., Matas, J., Obdr{\v{z}}{\'a}lek, {\v{S}}.: On evaluation of
  {6D} object pose estimation. In: ECCVW (2016)

\bibitem{hodan2018bop}
Hodan, T., Michel, F., Brachmann, E., Kehl, W., Buch, A.G., Kraft, D., Drost,
  B., Vidal, J., Ihrke, S., Zabulis, X., et~al.: {BOP}: Benchmark for {6D}
  object pose estimation. In: ECCV (2018)

\bibitem{hodan2015detection}
Hoda{\v n}, T., Zabulis, X., Lourakis, M., Obdr{\v z}{\'a}lek, {\v S}., Matas,
  J.: Detection and fine {3D} pose estimation of texture-less objects in
  {RGB-D} images. In: IROS (2015)

\bibitem{kehl2017ssd}
Kehl, W., Manhardt, F., Tombari, F., Ilic, S., Navab, N.: {SSD-6D}: Making
  {RGB}-based {3D} detection and {6D} pose estimation great again. In: ICCV
  (2017)

\bibitem{kehl2016deep}
Kehl, W., Milletari, F., Tombari, F., Ilic, S., Navab, N.: Deep learning of
  local {RGB-D} patches for {3D} object detection and {6D} pose estimation. In:
  ECCV (2016)

\bibitem{khoury17CGN}
Khoury, M., Zhou, Q.Y., Koltun, V.: Learning compact geometric features. In:
  ICCV (2017)

\bibitem{manhardt18}
Manhardt, F., Kehl, W., Navab, N., Tombari, F.: Deep model-based 6d pose
  refinement in rgb. In: ECCV (2018)

\bibitem{mellado2014super}
Mellado, N., Aiger, D., Mitra, N.J.: {Super4PCS Fast Global Pointcloud
  Registration via Smart Indexing}. In: Computer Graphics Forum. vol.~33, pp.
  205--215. Wiley Online Library (2014)

\bibitem{Mitash:2018aa}
Mitash, C., Boularias, A., Bekris, K.E.: {Robust 6D Object Pose Estimation with
  Stochastic Congruent Sets}. In: British Machine Vision Conference (BMVC)
  (2018)

\bibitem{mitash2017self}
Mitash, C., Bekris, K.E., Boularias, A.: A self-supervised learning system for
  object detection using physics simulation and multi-view pose estimation. In:
  Intelligent Robots and Systems (IROS), 2017 IEEE/RSJ International Conference
  on. pp. 545--551. IEEE (2017)

\bibitem{Mitash:2018ab}
Mitash, C., Boularias, A., Bekris, K.E.: Improving 6d pose estimation of
  objects in clutter via physics-aware monte carlo tree search. In: {IEEE}
  International Conference on Robotics and Automation (ICRA) (2018)

\bibitem{qi2017pointnet}
Qi, C.R., Su, H., Mo, K., Guibas, L.J.: Pointnet: Deep learning on point sets
  for 3d classification and segmentation. CVPR  (2017)

\bibitem{rad2017bb8}
Rad, M., Lepetit, V.: {BB8}: A scalable, accurate, robust to partial occlusion
  method for predicting the {3D} poses of challenging objects without using
  depth. In: ICCV (2017)

\bibitem{redmon2016you}
Redmon, J., Divvala, S., Girshick, R., Farhadi, A.: You only look once:
  Unified, real-time object detection. In: Proceedings of the IEEE conference
  on computer vision and pattern recognition. pp. 779--788 (2016)

\bibitem{ren2015faster}
Ren, S., He, K., Girshick, R., Sun, J.: Faster r-cnn: Towards real-time object
  detection with region proposal networks. In: Advances in Neural Information
  Processing Systems. pp. 91--99 (2015)

\bibitem{rennie2016dataset}
Rennie, C., Shome, R., Bekris, K.E., De~Souza, A.F.: A dataset for improved
  {RGBD}-based object detection and pose estimation for warehouse
  pick-and-place. Robotics and Automation Letters  (2016)

\bibitem{rethage18}
Rethage, D., Wald, J., Sturm, J., Navab, N., Tombari, F.: Fully-convolutional
  point networks for large-scale point clouds. In: ECCV (2018)

\bibitem{rusu09FPFH}
Rusu, R.B., Blodow, N., Beetz, M.: Fast point feature histograms (fpfh) for 3d
  registration. In: ICRA (2009)

\bibitem{tan2017looking}
Tan, D.J., Navab, N., Tombari, F.: Looking beyond the simple scenarios:
  combining learners and optimizers in 3d temporal tracking. IEEE Transactions
  on Visualization \& Computer Graphics (1), ~1--1 (2017)

\bibitem{tateno17CNNSLAM}
Tateno, K., Tombari, F., Laina, I., Navab, N.: Cnn-slam: Real-time dense
  monocular slam with learned depth prediction. In: CVPR (2017)

\bibitem{tejani2014latent}
Tejani, A., Tang, D., Kouskouridas, R., Kim, T.K.: Latent-class hough forests
  for {3D} object detection and pose estimation. In: ECCV (2014)

\bibitem{tekin18}
Tekin, B., Sinha, S.N., Fua, P.: Real-time seamless single shot 6d object pose
  prediction. In: CVPR (2018)

\bibitem{tombari10SHOT}
Tombari, F., Salti, S., Stefano, L.D.: Unique signatures of histograms for
  local surface description. In: ECCV (2010)

\bibitem{vidal2018sixd}
Vidal, J., Lin, C.Y., Mart{\'i}, R.: {6D} pose estimation using an improved
  method based on point pair features. In: ICCAR (2018)

\bibitem{xiang2017posecnn}
Xiang, Y., Schmidt, T., Narayanan, V., Fox, D.: {PoseCNN: A Convolutional
  Neural Network for 6D Object Pose Estimation in Cluttered Scenes}. arXiv
  preprint arXiv:1711.00199  (2017)

\bibitem{xiang18posecnn}
Xiang, Y., Schmidt, T., Narayanan, V., Fox, D.: Posecnn: A convolutional neural
  network for 6d object pose estimation in cluttered scenes. In: RSS (2018)

\bibitem{wang18DGCNN}
Yue~Wang, Yongbin~Sun, Z.L.S.E.S.M.M.B.J.M.S.: Dynamic graph cnn for learning
  on point clouds. arXiv preprint arXiv:1801.07829  (2018)

\bibitem{Princeton}
Zeng, A., Yu, K.T., Song, S., Suo, D., Walker~Jr, E., Rodriguez, A., Xiao, J.:
  Multi-view self-supervised deep learning for 6d pose estimation in the amazon
  picking challenge. In: {IEEE} International Conference on Robotics and
  Automation (ICRA) (2017)

\bibitem{zeng173DMatch}
Zeng, A., Song, S., Nie{\ss}ner, M., Fisher, M., Xiao, J., Funkhouser, T.:
  3dmatch: Learning local geometric descriptors from rgb-d reconstructions. In:
  CVPR (2017)

\end{thebibliography}

\end{document}